\documentclass{article}

\usepackage{arxiv}

\usepackage[utf8]{inputenc} 
\usepackage[T1]{fontenc}    
\usepackage{hyperref}       
\usepackage{url}            
\usepackage{booktabs}       
\usepackage{amsfonts}       
\usepackage{nicefrac}       
\usepackage{microtype}      
\usepackage{lipsum}		
\usepackage{graphicx}
\usepackage{doi}
\usepackage{amsmath}
\usepackage{biblatex} 
\title{ConTrip: Consensus Sentiment review Analysis and Platform ratings in a single score}

\author{{\hspace{1mm}Jos\'e Bonet} \\
	Department of Computational Biology\\
	IRB Barcelona\\
	Barcelona, Spain 08028 \\
	\texttt{jose.bonet@irbbarcelona.org} \\
	\And
	{\hspace{1mm}Jos\'e Bonet} \\
	Institute of Pure and Applied Mathematics\\
	Universitat Politècnica de València\\
	Valencia, Spain 46071\\
	\texttt{jbonet@mat.upv.es} \\
}

\renewcommand{\headeright}{}
\renewcommand{\undertitle}{}
\renewcommand{\shorttitle}{\textit{arXiv}}

\begin{document}
\maketitle

\begin{abstract}
	People unequivocally employ reviews to decide on purchasing an item or an experience on the internet. In that regard, the growing significance and number of opinions have led to the development of methods to assess their sentiment content automatically. However, it is not straightforward for the models to create a consensus value that embodies the agreement of the different reviews and differentiates across equal ratings for an item. Based on the approach proposed by Nguyen et al.\ \cite{nguyen2020fusing}, we derive a novel consensus value named ConTrip that merges their consensus score and the overall rating of a platform for an item. ConTrip lies in the rating range values, which makes it more interpretable while maintaining the ability to differentiate across equally rated experiences. ConTrip is implemented and freely available under MIT license at \href{https://github.com/pepebonet/contripscore}{github.com/pepebonet/contripscore}
\end{abstract}


\section{Introduction}

Customers in the world wide web do not rely solely on their opinion to acquire a product. Nowadays, ratings and reviews have proved their usefulness to influence consumers on their purchases \cite{amblee2011harnessing, keller2012word}. The reason is that these reviews facilitate decision-making, especially in experiences challenging to assess if not consumed as in hotels, restaurants, tours, or destinations \cite{liu2015makes}. Therefore, people's opinions put into words have significantly helped platforms such as Booking\footnote{booking.com}, Tripadvisor\footnote{tripadvisor.com}, Yelp\footnote{yelp.com}, and others to increase the flow of people setting their next experience through their platform. 

Given the relevance of these reviews in customer behavior, outstanding efforts have taken place to advance their automatic analysis \cite{fang2015sentiment}. As a consequence, the field of sentiment analysis has recently risen \cite{yadav2020sentiment}. Essentially, there are two basic approaches for automatic sentiment categorization --machine-learning (ML) \cite{hemmatian2019survey, zhang2018deep} and lexicon-based \cite{khoo2018lexicon}. Lexicon-based methods use handcrafted features and depend on a collection of lexical units and their sentiment orientation --sentiment lexicons. On the other hand, supervised machine learning methods (mainly used) identify word features that effectively distinguish between positive and negative sentiments.


Regardless of the technology employed, it is not straightforward to derive a measurement of agreement or disagreement between the reviews of a given product. Nguyen et al.\ \cite{nguyen2020fusing} proposed a score to measure the consensus of these reviews for a single item based on the sentiment analysis and rating of each critique. This new measurement allowed them to differentiate among ratings that were initially equal. That is, for instance, the ability to separate equivalent Tripadvisor ratings with their new inferred score. 

However, the consensus score embodies a value between 0 and 1, with values close to one representing total agreement and 0 disagreements in the reviews. This fact poses two significant complications: people can agree to good or bad reviews, and therefore one can not solely rely on the consensus value, and second a value between 0 and 1 is not easily interpretable for a public used to ratings between 1 and 5. 

Therefore, based on the work from Nguyen et al.\ \cite{nguyen2020fusing}, we propose an extension of their consensus value on our score named ConTrip. ConTrip is computed through a straightforward mathematical expression taking as input the consensus value of the reviews, for instance, as proposed by Nguyen et al.\ \cite{nguyen2020fusing} and the overall rating of the platform. As output, ConTrip encompasses a novel consensus value merging both inputs and rendering an interpretable value within the rating values. We argue that such a single value bringing together both sources of information could substitute the current overall rating of the platform to have a more truthful score of the experience to be purchased by the customer.

\section{Methods}
\label{sec:headings}

\subsection{ConTrip computation}

ConTrip employs a consensus value ($y \in [0, 1]$) of all the reviews as computed, for instance, from Nguyen et al.\ \cite{nguyen2020fusing} and the overall rating of an item ($x \in [1, 5]$). The consensus score embodies the sentiment analysis of every review and displays the degree of agreement across them. On the other hand, the overall rating represents the average of all the available ratings for an item. Through both values, ConTrip is denoted by $m$ and is computed as shown by Equation 1. 

\begin{equation}
      m:= {\underbrace{min(5, x + (y - 0.5) \cdot \alpha)}_{\text{Term 1}}}
      \ - 
      {\underbrace{\frac{(1 - y) \cdot x}{\beta}}_{\text{Term 2}}}
      \ - 
      {\underbrace{\frac{5 - x}{\delta}}_{\text{Term 3}}}
\end{equation}

The first term embodies a consensus penalty. If the consensus rating $y < 0.5$, the rating will decrease, while if $y > 0.5$, it will increase. A maximum value of 5 is enforced. The second term penalizes disagreements in a non-linear fashion --affecting disagreements more at higher ratings. The third and final term enforces differentiation across similar ratings. All equation terms are governed by a weight value ($\alpha$, $\beta$ and $\delta$). We fixed them heuristically to the following values: $\alpha = 0.5$, $\beta = 10$ and $\delta = 100$.

\subsection{ConTrip Scaling}

Given the nature of Equation 1 to compute ConTrip, score values lower than one can arise (minimum value 0.61). Therefore, to overcome difficulties in its applicability, we re-scale the ConTrip values $m$ from a [0.61, 5] range to a [1, 5] range employing the following equation: 

\begin{equation}
    (t_{max} - t_{min}) \cdot \frac{m - r_{min}}{r_{max} - r_{min}} + t_{min}
\end{equation}

where $t_{max}$ and $t_{min}$ denote the maximum and minimum of the range of the desired target scaling respectively (5 and 1). On the other hand, $r_{max}$ and $r_{min}$ denote the maximum and minimum of the range of the measurement.

\section{Results}

The objective of the work is to create an interpretable score that can differentiate across similarly rated items, all of this while merging the information of the sentiment analysis and the overall rating of an experience listed. This output score (ConTrip) would consist of a more informative overall rating.

To assess the proposed score, we first studied the behavior of ConTrip for 50 different values of any consensus score varying between 0 and 1 and for five overall ratings. For both cases, with and without scaling, Figures 1B and 1D show two characteristics: one, both the consensus score and the overall rating have significant contributions (For a Rating of 4, depending on the consensus values, ConTrip score ranges between 3.340 for consensus value of 0 and 4.240 for a consensus value of 1), and, two, as the rating increases, disagreement in the reviews (low consensus values) have a more powerful impact (A consensus of 0 drives a rating of 5 to have a ConTrip score of 4.25. In contrast, the same 0 consensus drives a rating of 2 only to 1.52). 

Secondly, we investigate what occurs when we fix six consensus values, and 41 different ratings are studied. Figures 1A and 1C (without and with scaling respectively) confirmed the previous trends and display the ability of ConTrip to differentiate between similarly rated items. Among all the Consensus-Rating pairs tested (246), the proposed ConTrip score achieved an overall differentiation of 97.6 \% (240 out of 246). 

\begin{figure*}[tpb]
\centering
\includegraphics[width=0.8\textwidth]{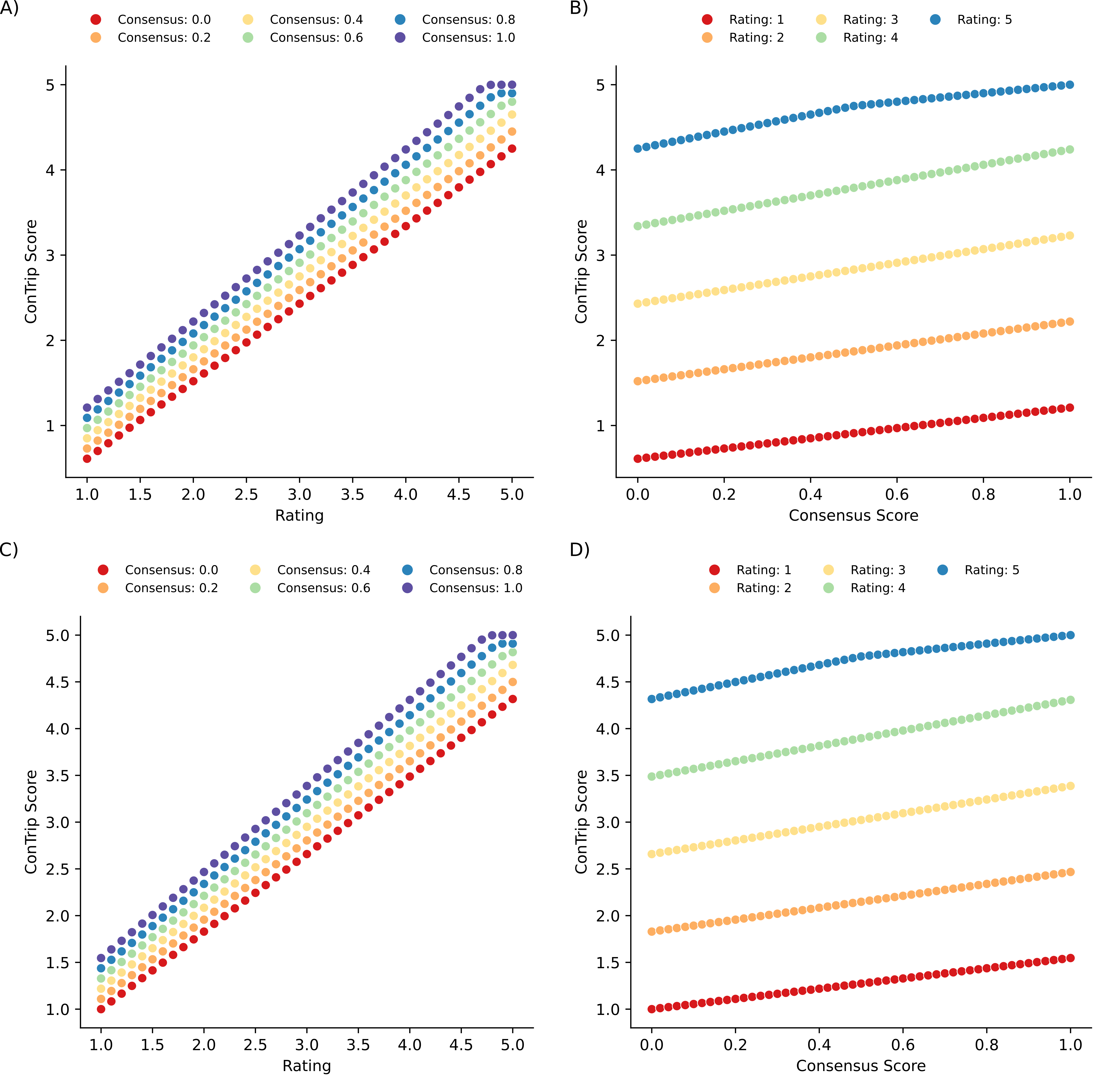}
\caption{Analysis of the ConTrip score for different consensus values from the sentiment analysis and different overall ratings for an item or experience from a platform. A, C) ConTrip measurements for a set of consensus values varying the platform rating. B,D) ConTrip measurements for a set of ratings varying the consensus values. A and B represent the ConTrip score without scaling and C and D with scaling. }
\label{F1}
\end{figure*}

\section{Discussion}

The current state of the sentiment analysis field is widely dominated by deep learning (DL) models achieving state-of-the-art performances for a wide range of tasks \cite{yadav2020sentiment}. The latest research shows the application of Transformers, Adversarial Learning and Graphical Neural Networks to Aspect-Based Sentiment Analysis \cite{liang2020aspect, peng2020knowing}, Sentiment Topic Extraction \cite{pergola2020disentangled}, and Emotion Caused Detection \cite{wei2020effective, xia2019rthn}, among other tasks. However, how the model output is used to give back information to the user may not be as straightforward. 

For that reason, works such as the one proposed by Nguyen et al.\ \cite{nguyen2020fusing}, despite not taking advantage of deep learning methodologies, are proven significant. On outtake of the publication was the ability of their consensus measurement to allow users to differentiate closely rated options. However, their proposed evaluation lacked interpretability as the range between 0 and 1 is not user-friendly, and values of 1, for instance, could be interpreted as both total positive or negative agreement. Therefore, and building upon their work, we developed ConTrip, a novel measurement that merges the consensus value of any sentiment analysis and the overall rating of an item listed. We showed that ConTrip does that while providing an interpretable and user-friendly measurement that can differentiate across closely rated experiences. 

We hypothesize that results from both lexicon-based and ML/DL approaches could be used as the input to compute ConTrip. Either as the centroid-based approach proposed by Nguyen et al.\ \cite{nguyen2020fusing} or as an average of the classification scores between Positive and Negative sentiment from a given method for a set of reviews. 

In summary, ConTrip poses an initial milestone to substitute overall rating scores with a more informative measurement. User experience is crucial, and including an additional value or a set of numbers may pose risks for the website conversion rates. Therefore, ConTrip embodies all the information in a single value and could be directly implemented. As DL continues to evolve and the data extracted from reviews are more accurate and valuable, their possible impact on the item rating will be higher. Therefore, their inclusion on the overall rating will be even more compelling. ConTrip is just an initial step in that direction.

\section{Acknowledgements}
The authors want to thank Nuria Agell, Albert Armisen and Jennifer Nguyen for the discussion about their publication that made this possible.

\printbibliography

@article{nguyen2020fusing,
  title={Fusing hotel ratings and reviews with hesitant terms and consensus measures},
  author={Nguyen, Jennifer and Montserrat-Adell, Jordi and Agell, N{\'u}ria and S{\'a}nchez, Monica and Ruiz, Francisco J},
  journal={Neural Computing and Applications},
  volume={32},
  number={19},
  pages={15301--15311},
  year={2020},
  publisher={Springer}
}

@article{khoo2018lexicon,
  title={Lexicon-based sentiment analysis: Comparative evaluation of six sentiment lexicons},
  author={Khoo, Christopher SG and Johnkhan, Sathik Basha},
  journal={Journal of Information Science},
  volume={44},
  number={4},
  pages={491--511},
  year={2018},
  publisher={SAGE Publications Sage UK: London, England}
}

@article{yadav2020sentiment,
  title={Sentiment analysis using deep learning architectures: a review},
  author={Yadav, Ashima and Vishwakarma, Dinesh Kumar},
  journal={Artificial Intelligence Review},
  volume={53},
  number={6},
  pages={4335--4385},
  year={2020},
  publisher={Springer}
}

@article{zhang2018deep,
  title={Deep learning for sentiment analysis: A survey},
  author={Zhang, Lei and Wang, Shuai and Liu, Bing},
  journal={Wiley Interdisciplinary Reviews: Data Mining and Knowledge Discovery},
  volume={8},
  number={4},
  pages={e1253},
  year={2018},
  publisher={Wiley Online Library}
}

@article{hemmatian2019survey,
  title={A survey on classification techniques for opinion mining and sentiment analysis},
  author={Hemmatian, Fatemeh and Sohrabi, Mohammad Karim},
  journal={Artificial Intelligence Review},
  volume={52},
  number={3},
  pages={1495--1545},
  year={2019},
  publisher={Springer}
}

@article{fang2015sentiment,
  title={Sentiment analysis using product review data},
  author={Fang, Xing and Zhan, Justin},
  journal={Journal of Big Data},
  volume={2},
  number={1},
  pages={1--14},
  year={2015},
  publisher={SpringerOpen}
}

@article{liu2015makes,
  title={What makes a useful online review? Implication for travel product websites},
  author={Liu, Zhiwei and Park, Sangwon},
  journal={Tourism management},
  volume={47},
  pages={140--151},
  year={2015},
  publisher={Elsevier}
}

@article{amblee2011harnessing,
  title={Harnessing the influence of social proof in online shopping: The effect of electronic word of mouth on sales of digital microproducts},
  author={Amblee, Naveen and Bui, Tung},
  journal={International journal of electronic commerce},
  volume={16},
  number={2},
  pages={91--114},
  year={2011},
  publisher={Taylor \& Francis}
}

@article{keller2012word,
  title={Word-of-mouth advocacy: A new key to advertising effectiveness},
  author={Keller, Ed and Fay, Brad},
  journal={Journal of Advertising Research},
  volume={52},
  number={4},
  pages={459--464},
  year={2012},
  publisher={Journal of Advertising Research}
}

@inproceedings{peng2020knowing,
  title={Knowing what, how and why: A near complete solution for aspect-based sentiment analysis},
  author={Peng, Haiyun and Xu, Lu and Bing, Lidong and Huang, Fei and Lu, Wei and Si, Luo},
  booktitle={Proceedings of the AAAI Conference on Artificial Intelligence},
  volume={34},
  number={05},
  pages={8600--8607},
  year={2020}
}

@inproceedings{liang2020aspect,
  title={Aspect-invariant Sentiment Features Learning: Adversarial Multi-task Learning for Aspect-based Sentiment Analysis},
  author={Liang, Bin and Yin, Rongdi and Gui, Lin and Du, Jiachen and He, Yulan and Xu, Ruifeng},
  booktitle={Proceedings of the 29th ACM International Conference on Information \& Knowledge Management},
  pages={825--834},
  year={2020}
}

@article{pergola2020disentangled,
  title={A Disentangled Adversarial Neural Topic Model for Separating Opinions from Plots in User Reviews},
  author={Pergola, Gabriele and Gui, Lin and He, Yulan},
  journal={arXiv preprint arXiv:2010.11384},
  year={2020}
}

@article{xia2019rthn,
  title={RTHN: A rnn-transformer hierarchical network for emotion cause extraction},
  author={Xia, Rui and Zhang, Mengran and Ding, Zixiang},
  journal={arXiv preprint arXiv:1906.01236},
  year={2019}
}

@inproceedings{wei2020effective,
  title={Effective inter-clause modeling for end-to-end emotion-cause pair extraction},
  author={Wei, Penghui and Zhao, Jiahao and Mao, Wenji},
  booktitle={Proceedings of the 58th Annual Meeting of the Association for Computational Linguistics},
  pages={3171--3181},
  year={2020}
}

\end{document}